\documentclass[11pt,a4paper]{article}
\usepackage[hyperref]{emnlp2018}
\usepackage{times}
\usepackage{latexsym}
\usepackage{booktabs}
\usepackage{enumitem}

\usepackage{adjustbox}
\usepackage{calc}

\usepackage{tikz}
\usepackage{pgfplots}
\pgfplotsset{compat=newest}

\aclfinalcopy 

\newcommand\hate{\textsc{hate}}
\newcommand\nohate{\textsc{noHate}}
\newcommand\idk{\textsc{skip}}
\newcommand\relation{\textsc{relation}}

\newcommand{\mli}[1]{\mathit{#1}}

\newcounter{examples}
\newcounter{subexamples}
\newlist{example}{enumerate}{10}
\setlist[example]{label=(\theexamples), align=left, leftmargin=2.75em, resume,after=\setcounter{subexamples}{1}}
\let\olditem\item
\newcommand{\exitem}{\olditem \stepcounter{examples}}	
\newcommand{\sexitem}{\olditem \stepcounter{subexamples}}
\renewcommand{\item}{\olditem}

\newlist{inline}{enumerate*}{1}
\setlist*[inline]{label=\textit{\alph*)}, itemjoin={{, }}, itemjoin*={{, and }}}

\title{Hate Speech Dataset from a White Supremacy Forum}  
\author{Ona de Gibert\qquad Naiara Perez\qquad Aitor Garc\'ia-Pablos\qquad Montse Cuadros\\
	HSLT Group at Vicomtech, Donostia/San Sebasti\'an, Spain \\
	{\tt\{odegibert,nperez,agarciap,mcuadros\}@vicomtech.org}}

\date{}

\begin{document}

\maketitle


\begin{abstract}
Hate speech is commonly defined as any communication that disparages a target group of people based on some characteristic such as race, colour, ethnicity, gender, sexual orientation, nationality, religion, or other characteristic.
Due to the massive rise of user-generated web content on social media, the amount of hate speech is also steadily increasing. Over the past years, interest in online hate speech detection and, particularly, the automation of this task has continuously grown, along with the societal impact of the phenomenon.
This paper describes a hate speech dataset composed of thousands of sentences manually labelled as containing hate speech or not. The sentences have been extracted from Stormfront, a white supremacist forum. A custom annotation tool has been developed to carry out the manual labelling task which, among other things, allows the annotators to choose whether to read the context of a sentence before labelling it. The paper also provides a thoughtful qualitative and quantitative study of the resulting dataset and several baseline experiments with different classification models. 
The dataset is publicly available.
\end{abstract}


\section{Introduction}
\label{sec:intro}

The rapid growth of content in social networks such as Facebook, Twitter and blogs, makes it impossible to monitor what is being said. The increase of cyberbullying and cyberterrorism, and the use of hate on the Internet, make the identification of hate in the web an essential ingredient for anti-bullying policies of social media, as Facebook's CEO Mark Zuckerberg recently acknowledged\footnote{https://www.washingtonpost.com/news/the-switch/wp/2018/04/10/transcript-of-mark-zuckerbergs-senate-hearing/}. This paper releases a new dataset of hate speech to further investigate the problem.

Although there is no universal definition for \textit{hate speech}, the most accepted definition is provided by \citet{nockleby2000hate}: ``any communication that disparages a target group of people based on some characteristic such as race, colour, ethnicity, gender, sexual orientation, nationality, religion, or other characteristic''. Consider the following\footnote{The examples in this work may contain offensive language. They have been taken from actual web data and by no means reflect the authors' opinion.}:
\begin{example}
\exitem ``God bless them all, to hell with the blacks''
\end{example}
This sentence clearly contains hate speech against a target group because of their skin colour. However, the identification of hate speech is often not so straightforward. Besides defining hate speech as a verbal abuse directed to a group of people because of specific characteristics, other definitions of hate speech in previous studies care to include the speaker's determination to inflect harm \cite{Davidson2017}.

In all, there seems to be a pattern shared by most of the literature consulted \citep{nockleby2000hate,Djuric2015,Gitari2015,Nobata2016,Silva2016,Davidson2017}, which would define hate speech as
\begin{inline}
\item a deliberate attack
\item directed towards a specific group of people
\item motivated by actual or perceived aspects that form the group's identity.
\end{inline}

This paper presents the first public dataset of hate speech annotated on Internet forum posts in English at sentence-level. The dataset is publicly available in GitHub\footnote{https://github.com/aitor-garcia-p/hate-speech-dataset}. The source forum is Stormfront\footnote{www.stormfront.org}, the largest online community of white nationalists, characterised by pseudo-rational discussions of race \cite{meddaugh2009}, which include different degrees of offensiveness. Stormfront is known as the first hate website \cite{schafer2002}.

The rest of the paper is structured as follows: Section \ref{sec:related} describes the related work and contextualises the work presented in the paper; Section \ref{sec:hate} introduces the task of generating a manually labelled hate speech dataset; this includes the design of the annotation guidelines, the resulting criteria, the  inter-annotator agreement and a quantitative description of the resulting dataset; next, Section \ref{sec:experiments} presents several baseline experiments with different classification models using the labelled data; finally, Section \ref{sec:disc} provides a brief discussion about the difficulties and nuances of hate speech detection, and Section \ref{sec:conc} summarises the conclusions and future work.


\section{Related Work}
\label{sec:related}
Research on hate speech has increased in the last years. The conducted studies are diverse and work on different datasets; there is no official corpus for the task, so usually authors collect and label their own data. For this reason, there exist few publicly available resources for hate speech detection. 

Hatebase\footnote{https://www.hatebase.org/} is the an online repository of structured, multilingual, usage-based hate speech. Its vocabulary is classified into eight categories: archaic, class, disability, ethnicity, gender, nationality, religion, and sexual orientation. Some studies make use of Hatebase to build a classifier for hate speech \cite{Davidson2017,Serra2017, Nobata2016}. However, \citet{Saleem2017} prove that keyword-based approaches succeed at identifying the topic but fail to distinguish hateful sentences from clean ones, as the same vocabulary is shared by the hateful and target community, although with different intentions. 

Kaggle's Toxic Comment Classification Challenge dataset\footnote{https://www.kaggle.com/c/jigsaw-toxic-comment-classification-challenge/data} consists of 150k Wikipedia comments annotated for toxic behaviour. \citet{WaseemHovy2016} published a collection of 16k tweets classified into racist, sexist or neither. \citet{Sharma2018} collected a set of 9k tweets containing harmful speech and they manually annotated them based on their degree of hateful intent. They describe three different classes of hate speech. The definition on which this paper is based overlaps mostly with their Class I, described as speech 
\begin{inline}
\item that incites violent actions
\item directed at a particular group
\item with the intention of conveying hurting sentiments.
\end{inline}

Google and Jigsaw developed a tool called Perspective\footnote{https://www.perspectiveapi.com} that measures the ``toxicity'' of comments. The tool is published as an API and gives a toxicity score between 0 and 100 using a machine learning model. Such model has been trained on thousands of comments manually labelled by a team of people\footnote{https://www.nytimes.com/2017/02/23/technology/google-jigsaw-monitor-toxic-online-comments.html}; to our knowledge, the resulting dataset is not publicly available.

The detection of hate speech has been tackled in three main different ways. Some studies focus on subtypes of hate speech. This is the case of \citet{Warner2012}, who focus on the identification of anti-Semitic posts versus any other form of hate speech. Also in this line, \citet{Kwok2013} target anti-black hate speech. \citet{Badjatiya2017,gamback2017} study the detection of racist and sexist tweets using deep learning.

Other proposals focus on the annotation of hate speech as opposed to texts containing derogatory or offensive language \cite{Davidson2017,Malmasi2017,Malmasi2018,Watanabe2018}. They build multi-class classifiers with the categories ``hate'', ``offensive'', and ``clean''. 

Finally, some studies focus on the annotation of hate speech versus clean comments that do not contain hate speech \cite{Nobata2016,Burnap2015,Djuric2015}. \citet{Gitari2015} follow this approach but further classify the hateful comments into two categories: ``weak'' and ``strong'' hate. \citet{DelVigna2017} conduct a similar study for Italian.

In all, experts conclude that annotation of hate speech is a difficult task, mainly because of the data annotation process. \citet{Waseem2016} conducted a study on the influence of annotator knowledge of hate speech on classifiers for hate speech. \citet{ross2017} also studied the reliability of hate speech annotations and acknowledge the importance of having detailed instructions for the annotation of hate speech available.

This paper aims to tackle the inherent subjectivity and difficulty of labelling hate speech by following strict guidelines. The approach presented in this paper follows \cite{Nobata2016,Burnap2015,Djuric2015} (i.e., ``hateful'' versus ``clean''). Furthermore, the annotation has been performed at sentence level as opposed to full-comment annotation, with the possibility to access the original complete post for each sentence. To our knowledge, this is the first work that releases a manually labelled hate speech dataset annotated at sentence level in English posts from a white supremacy forum.


\section{Hate Speech Dataset}
\label{sec:hate}

This paper presents the first dataset of textual hate speech annotated at sentence-level. Sentence-level annotation allows to work with the minimum unit containing hate speech and reduce noise introduced by other sentences that are clean. 

A total number of 10,568 sentences have been extracted from Stormfront and classified as conveying hate speech or not, and into two other auxiliary classes, as per the guidelines described in Section \ref{ssec:guidelines}. In addition, the following information is also given for each sentence: a post identifier and the sentence's position in the post, a user identifier, a sub-forum identifier\footnote{All the identifiers provided are fake placeholders that facilitate understanding relations between sentences, Stormfront users, etc., but do not point back to the original source.}. This information makes it possible re-build the conversations these sentences belong to. Furthermore, the number of previous posts the annotator had to read before making a decision over the category of the sentence is also given. 

\subsection{Data extraction and processing}
\label{ssec:data_extraction}

The content was extracted from Stormfront using web-scraping techniques and was dumped into a database arranged by sub-forums and conversation threads \citep{Figea2016}. The extracted forum content was published between 2002 and 2017. The process of preparing the candidate content to be annotated was the following:

\begin{enumerate}[noitemsep]
\item A subset of 22 sub-forums covering diverse topics and nationalities was random-sampled to gather individual posts uniformly distributed among sub-forums and users.
\item The sampled posts were filtered using an automatic language detector\footnote{https://github.com/shuyo/language-detection/blob/wiki/ProjectHome.md} to discard non-English texts.
\item The resulting posts were segmented into sentences with ixa-pipes \cite{Agerri2014}.
\item The sentences were grouped forming batches of 500 complete posts ($\sim$ 1,000 sentences per batch).
\end{enumerate}

The manual annotation task was divided into batches to control the process. During the annotation of the first two batches, the annotation procedure and guidelines were progressively refined and adapted. In total, 10,568 sentences contained in 10 batches have been manually annotated.

A post-processing step was performed to filter excessively long or short sentences. The cleansing process removed sentences shorter than 3 words or longer than 50 words. In total, 652 sentences were discarded, which represent 6.17\% of the original dataset. The resulting dataset (henceforth referred to as the ``clean'' dataset) is the one described in Section \ref{ssec:stats} and used for experimentation in Section \ref{sec:experiments}. The release of the dataset will contain both the raw sentences without any post-processing and the clean version, both annotated.


\subsection{Annotation guidelines}
\label{ssec:guidelines}

\citet{Schmidt2017} acknowledge that the procedure for hate speech annotation is fairly vague in previous studies, which translates into low agreement scores. In this study, all the annotators together created and discussed the guidelines to ensure all participants had the same understanding of hate speech. The final guidelines consider 4 types of sentences: 

\subsubsection{\hate}
\label{sssec:hate}

Sentences in this category contain hate speech. Hate speech is a
\begin{enumerate}[label=\alph*), noitemsep]
\item \label{it:attack} deliberate attack
\item \label{it:group} directed towards a specific group of people
\item \label{it:identity} motivated by aspects of the group's identity.
\end{enumerate}
The three premises must be true for a sentence to be categorized as \hate. Consider the following examples:
\begin{example}
\exitem \label{it:apes} ``Poor white kids being forced to treat apes and parasites as their equals.'' 
\exitem \label{it:islam} ``Islam is a false religion however unlike some other false religions it is crude and appeals to crude people such as arabs.''
\end{example}
In \ref{it:apes}, the speaker uses ``apes'' and ``parasites'' to refer to children of dark skin and implies they are not equal to ``white kids''. That is, it is an attack to the group composed of children of dark skin based on an identifying characteristic, namely, their skin colour. Thus, all the premises are true and \ref{it:apes} is a valid example of \hate. Example \ref{it:islam} brands all people of Arab origin as crude. That is, it attacks the group composed of Arab people based on their origin. Thus, all the premises are true and \ref{it:islam} is a valid example of \hate.

\subsubsection{\nohate}
\label{sssec:nohate}

This label is used to categorise sentences that do not convey hate speech per the established definition. Consider the following examples:
\begin{example}
\exitem \label{it:ns} ``Where can I find NS speeches and music, also historical, in mp3 format for free download on the net.``
\exitem \label{it:chris} ``I know of Chris Rock and subsequently have hated him for a long time.''
\end{example}
Example \ref{it:ns} mentions National Socialism (``NS''), but the user is just interested in documentation about it. Therefore, the sentence itself is not an attack, i.e., premise \ref{it:attack} is not true, despite the sound assumption that the speaker forms part of a hating community. Thus, \ref{it:ns} is not a valid instance of \hate. Example \ref{it:chris} is directed towards an individual; thus, premise \ref{it:group} is false and the sentence is not a valid example of \hate, despite the sound assumption that the attack towards the individual is based on his skin colour. 

Finally, it must be emphasized that the presence of pejorative language in a sentence cannot systematically be considered sufficient evidence to confirm the existence of hate speech. The use of ``fag'' in the following sentence:
\begin{example}
\exitem \label{it:fags} ``Two black fag's holding hands.''
\end{example}
\noindent cannot be said to be a deliberate attack, taken without any more context, despite it likely being offensive. Therefore, it cannot be considered \hate.

\subsubsection{\relation}
\label{sssec:relation}

When \ref{it:fags} (repeated as \ref{it:off1}) is read in context:
\begin{example}[label=(\theexamples.\thesubexamples), noitemsep]
\sexitem \label{it:off1} ``Two black fag's holding hands.''
\sexitem \label{it:off2} ``That's Great!''
\sexitem \label{it:off3} ``That's 2 blacks won't be having kids.'' \stepcounter{examples}
\end{example}
\noindent it clearly conveys hate speech. The author is celebrating that two people belonging to the black minority will not be having children, which is a deliberate attack on a group of people based on an identifying characteristic. The annotation at sentence-level fails to discern that there exists hate speech in this example. The label \relation\ is for specific cases such as this, where the sentences in a post do not contain hate speech on their own, but the combination of several sentences does. Consider another example:
\begin{example}[label=(\theexamples.\thesubexamples), noitemsep]
\sexitem \label{it:rel1} ``Probably the most disgusting thing I've seen in the last year.''
\sexitem \label{it:rel2} ``She looks like she has some African blood in her, or maybe it's just the makeup.''
\sexitem \label{it:rel3} ``This is just so wrong.'' \stepcounter{examples}
\end{example}
Each sentence in isolation does not convey hate speech: in \ref{it:rel1} and \ref{it:rel3}, a negative attitude is perceived, but it is unknown whether it is targeted towards a group of people; in \ref{it:rel2}, there is no hint of an attack, not even of a negative attitude. However, the three sentences together suggest that having ``African blood'' makes a situation (whatever ``this'' refers to) disgusting, which constitutes hate speech according to the definition proposed.

The label \relation\ is given separately to all the sentences that need each other to be understood as hate speech. That is, consecutive sentences with this label convey hate speech but depend on each other to be correctly interpreted. 

\subsubsection{\idk}
\label{sssec:idk}

Sentences that are not written in English or that do not contain information as to be classified into \hate\ or \nohate\ are given this label.
\begin{example}
\exitem \label{it:nor} ``Myndighetene vurderer nå om de skal få permanent oppholdstillatelse.''
\exitem \label{it:yt} ``YouTube - Broadcast Yourself.''
\end{example}
Example \ref{it:nor} is in Norwegian and \ref{it:yt} is irrelevant both for \hate\ and \nohate.

\subsection{Annotation procedure}
\label{ssec:procedure}

In order to develop the annotation guidelines, a draft was first written based on previous similar work. Three of the authors annotated a 1,144-sentence batch of the dataset following the draft, containing only the categories \hate, \nohate\ and \idk. Then, they discussed the annotations and modified the draft accordingly, which resulted in the guidelines presented in the previous section, including the \relation\ category. Finally, a different batch of 1,018 sentences was annotated by the same three authors adhering to the new guidelines in order to calculate the inter-annotator agreement. 

Table \ref{tab:agreement} shows the agreements obtained in terms of the average percent agreement ($avg\ \%$), average Cohen's kappa coefficient \cite{Cohen1960} ($avg\ k$), and Fleiss' kappa coefficient \cite{fleiss1971measuring} ($fleiss$). The number of annotated sentences (\# sent) and the number of categories to label (\# cat) are also given for each batch. The results are in line with similar works \cite{Nobata2016, Warner2012}.

\begin{table}[!htbp]
\centering
\begin{tabular}{@{}cccccc@{}}
\toprule
   & \# sent &\# cat & $avg\ \%$ & $avg\ k$ & $fleiss$ \\ \midrule
1 & 1,144   & 3    & 91.03    & 0.614    & 0.607    \\
2 & 1,018   & 4    & 90.97    & 0.627    & 0.632    \\ \bottomrule
\end{tabular}
\caption{Inter-annotator agreements on batches 1 and 2}
\label{tab:agreement}
\end{table}

All the annotation work was carried out using a web-based tool developed by the authors for this purpose. The tool displays all the sentences belonging to the same post at the same time, giving the annotator a better understanding of the post's author's intention. If the complete post is deemed insufficient by the annotator to categorize a sentence, the tool can show previous posts to which the problematic post is answering, on demand, up to the first post in the thread and its title. This consumption of context is registered automatically by the tool for further treatment of the collected data.

As stated by other studies, context appears to be of great importance when annotating hate speech \cite{Watanabe2018}. \citet{Schmidt2017} acknowledge that whether a message contains hate speech or not can depend solely on the context, and thus encourage the inclusion of extra-linguistic features for annotation of hate speech. Moreover, \citet{Sharma2018} claim that context is essential to understand the speaker's intention. 


\subsection{Dataset statistics}
\label{ssec:stats}

This section provides a quantitative description and statistical analysis of the clean dataset published. Table \ref{tab:categories} shows the distribution of the sentences over categories. The dataset is unbalanced as there exist many more sentences not conveying hate speech than `hateful'' ones.

\begin{table}[!htbp]
\centering
\begin{tabular}{@{}rrr}
\toprule
 \multicolumn{1}{c}{Assigned label} & \multicolumn{1}{c}{\# sent} & \multicolumn{1}{c}{\%} \\ \midrule
\hate     & 1,119  & 11.29  \\
\nohate   & 8,537  & 86.09    \\
\relation & 168    & 1.69      \\
\idk      & 92     & 0.93   \\
total     & 9,916  & 100.00   \\ \bottomrule
\end{tabular}
\caption{Distribution of sentences over categories in the clean dataset}
\label{tab:categories}
\end{table}

Table \ref{tab:context_nocontext} refers to the subset of sentences that have required reading additional context (i.e. previous comments to the one being annotated) to make an informed decision by the human annotators. The category \hate\ is the one that requires more context, usually due to the use of slang unknown to the annotator or because the annotator needed to find out the actual target of an offensive mention.

\begin{table}[!htbp]
\centering
\begin{tabular}{@{}rrrr@{}}
\toprule
                & \multicolumn{1}{c}{Context used} & \multicolumn{1}{c}{No context used} \\ \midrule
\hate       & 22.70              & 77.30           \\
\nohate & 8.00 & 92.00 \\ \bottomrule
\end{tabular}
\caption{Percentage of sentences for which the human annotators asked for additional context} 
\label{tab:context_nocontext}
\end{table}

The remaining of the section focuses only on the subset of the dataset composed of the categories \hate\ and \nohate, which are the core of this work. Table \ref{tab:statistics} shows the size of said subset, along with the average sentence length for each class, their word counts and their vocabulary sizes. 

\begin{table}[!htbp]
\centering
\begin{tabular}{@{}rrrr@{}}
\toprule
                & \multicolumn{1}{c}{\hate} & \multicolumn{1}{c}{\nohate} \\ \midrule
sentences       & 1,119              & 8,537            \\
sentence length & 20.39 $\pm$ 9.46 & 15.15 $\pm$ 9.16 \\
word count      & 24,867           & 144,353           \\
vocabulary      & 4,148            & 13,154            \\ \bottomrule
\end{tabular}
\caption{Size of the categories \hate\ and \nohate\ in the clean dataset}
\label{tab:statistics}
\end{table}

Regarding the distribution of sentences over Stormfront accounts, the dataset is balanced as there is no account that contributes notably more than any other:
the average percentage of sentences is of 0.50 $\pm$ 0.42 per account, the total amount of accounts in the dataset being 2,723. 
The sub-forums that contain more \hate\ belong to the category of news, discussion of views, politics, philosophy, as well as to specific countries (i.e., Ireland, Britain, and Canada). In contrast, the sub-forums that contain more \nohate\ sentences are about education and homeschooling, gatherings, and youth issues.

In order to obtain a more qualitative insight of the dataset, a \hate\ score ($\textsc{hs}$) has been calculated based on the Pointwise Mutual Information (PMI) value for each word towards the categories \hate\ and \nohate. PMI allows calculating the correlation of each word with respect to each category. The difference of the PMI value of a word $w$ and the category \hate\ and the PMI of the same word $w$ and the category \nohate\ results in the \hate\ score of $w$, as shown in Formula \ref{eq:pmi_hate_nohate}.

\begin{equation}
\textsc{hs}({\small w}) = \mli{PMI}({\small w,\hate}) - \mli{PMI}({\small w,\nohate})
\label{eq:pmi_hate_nohate}
\end{equation}

Intuitively, this score is a simple way of capturing whether the presence of a word in a \hate\ context occurs significantly more often than in a \nohate\ context.
Table \ref{tab:pmi} shows the 15 most and least hateful words: the more positive a \hate\ score, the more hateful a word, and vice versa.

\begin {table}[!htbp]
\centering
\begin{tabular}{@{}rr|rr@{}}
\toprule
 & \multicolumn{1}{c|}{\textsc{hs}} &  & \multicolumn{1}{c}{\textsc{hs}}  \\ \midrule
ape         & 6.81  & pm          & -3.38 \\
scum        & 6.25  & group       & -3.34 \\
savages     & 5.73  & week        & -3.13 \\
filthy      & 5.73  & idea        & -2.70 \\
mud         & 5.31  & thread      & -2.68 \\
homosexuals & 5.31  & german      & -2.67 \\
filth       & 5.19  & videos      & -2.67 \\
apes        & 5.05  & night       & -2.63 \\
beasts      & 5.05  & happy       & -2.63 \\
homosexual  & 5.05  & join        & -2.63 \\
threat      & 5.05  & pictures    & -2.60 \\
monkey      & 5.05  & eyes        & -2.54 \\
libtard     & 5.05  & french      & -2.52 \\
coon        & 5.05  & information & -2.44 \\
niglet      & 4.73  & band        & -2.44 \\ \bottomrule
\end{tabular}
\caption {Most (positive \textsc{hs}) and least (negative \textsc{hs}) hateful words} 
\label{tab:pmi} 
\end{table}

The results show that the most hateful words are derogatory or refer to targeted groups of hate speech. On the other hand, the least hateful words are neutral in this regard and belong to the semantic fields of Internet, or temporal expressions, among others. This shows that the vocabulary is discernible by category, which in turn suggests that the annotation and guidelines are sound.

Performing the same calculation with bi-grams yields expressions such as ``gene pool'', ``race traitor'', and ``white guilt'' for the most hateful category, which appear to be concepts related to race issues. The less hateful terms are expressions such as ``white power'', ``white nationalism'' and ``pro white'', which clearly state the right-wing extremist politics of the forum users.

Finally, the dataset has been contrasted against the English vocabulary in Hatebase. 9.28\% of \hate\ vocabulary overlaps with Hatebase, a higher percentage than for \nohate\ vocabulary, of which 6.57\% of the words can be found in Hatebase. In Table \ref{tab:hatebase}, the distribution of \hate\ vocabulary is shown over Hatebase's 8 categories. Although some percentages are not high, all 8 categories are present in the corpus.
Most of the \hate\ words from the dataset belong to ethnicity, followed by gender. This is in agreement with \citet{Silva2016}, who conducted a study to analyse the targets of hate in social networks and showed that hate based on race was the most common.

\begin{table}[!htbp]
\centering
\resizebox{\columnwidth}{!}{%
\begin{tabular}{lrc}
\toprule
\multicolumn{1}{c}{category} & \multicolumn{1}{c}{\%} & \multicolumn{1}{c}{examples}\\ \midrule
archaic            & 2.46					   & div, wigger       \\
ethnicity          & 41.63                     & coon, paki        \\
nationality        & 7.03                      & guinea, leprechaun \\
religion           & 1.34                      & holohoax, prod     \\
gender             & 36.05                     & bird, dyke         \\
sexual orientation & 2.34                      & fag, queer         \\
disability         & 2.01                      & mongol, retarded   \\
social class       & 7.14  					   & slag, trash     \\ \midrule
total              & 100.00 				   &  \\ \bottomrule
\end{tabular}
}
\caption{Distribution of \hate\ vocabulary over Hatebase categories}
\label{tab:hatebase} 
\end{table}


\section{Experiments}
\label{sec:experiments}

In order to further inspect the resulting dataset and to check the validity of the annotations (i.e. whether the two annotated classes are separable based solely on the text of the labelled instances) a set of baseline experiments have been conducted. These experiments do not exploit any external resource such as lexicons, heuristics or rules. The experiments just use the provided dataset and well-known approaches from the literature to provide a baseline for further research and improvement in the future.

\subsection{Experimental setting}
\label{sec:experimental_setting}

The experiments are based on a balanced subset of labelled sentences. All the sentences labelled as \hate\ have been collected, and an equivalent number of \nohate\ sentences have been randomly sampled, summing up 2k labelled sentences. From this amount, the 80\% has been used for training and the remaining 20\% for testing.

The evaluated algorithms are the following:

\begin{itemize}[noitemsep]
\item Support Vector Machines (SVM) \citep{hearst1998support} over Bag-of-Words vectors. Word-count-based vectors have been computed and fed into a Python Scikit-learn LinearSVM\footnote{http://scikit-learn.org/stable/modules/svm.html} classifier to separate \hate\ and \nohate\ instances.
\item Convolutional Neural Networks (CNN), as described in \citep{Kim2014}. The implementation is a simplified version using a single input channel of randomly initialized word embeddings\footnote{https://github.com/dennybritz/cnn-text-classification-tf}.
\item Recurrent Neural Networks with Long Short-term Memories (LSTM) \citep{hochreiter1997long}. A LSTM layer of size 128 over word embeddings of size 300.
\end{itemize}

All the hyperparameters are left to the usual values reported in the literature \citep{greff2017lstm}. No hyperparameter tuning has been performed. A more comprehensive experimentation and research has been left for future work.


\subsection{Results}
\label{sec:results}

The baseline experiments include a majority class baseline showing the balance between the two classes in the test set. The results are given in terms of accuracy for \hate and \nohate\ individually, and the overall accuracy, calculated according to the equations \ref{eq:tp_hate}, \ref{eq:tp_nohate} and \ref{eq:tp_all}, where $\mli{TP}$ are the true positives and $\mli{FP}$ are the false positives.
\begin{equation}
Acc_{\small\hate} = \frac{\mli{TP}_{\small\hate}}{\mli{TP}_{\small\hate}+\mli{FP}_{\small\hate}}
\label{eq:tp_hate}
\end{equation}

\begin{equation}
Acc_{\small\nohate} = \frac{\mli{TP}_{\small\nohate}}{\mli{TP}_{\small\nohate}+\mli{FP}_{\small\nohate}}
\label{eq:tp_nohate}
\end{equation}

\begin{equation}
Acc_{\small\textsc{all}} = \frac{\mli{TP}_{\small\textsc{all}}}{\mli{TP}_{\small\textsc{all}}+\mli{FP}_{\small\textsc{all}}}
\label{eq:tp_all}
\end{equation}

We show the accuracy for the both complementary classes instead of the precision-recall of a single class to highlight the performance of the classifiers for the both classes individually.
Table \ref{tab:results_without_context} shows the results of using only sentences that did \textit{not} require additional context to be labelled, while Table \ref{tab:results_with_context} shows the results of including those sentences that required additional context. Not surprisingly, the results are lower when including sentences that required additional context. If a human annotator required additional information to make a decision, it is to expect that an automatic classifier would not have enough information or would have a harder time making a correct prediction. The results also show that \nohate\ sentences are more accurately classified than \hate\ sentences. Overall, the LSTM-based classifier obtains better results, but even the simple SVM using bag-of-words vectors is capable of discriminating the classes reasonably well.

\begin{table}[htbp]
	\centering
    \begin{tabular}{@{}rccc@{}}
    	\toprule
		& $Acc_{\small\hate}$ & $Acc_{\small\nohate}$ & $Acc_{\small\textsc{all}}$ \\ \midrule
        Majority & n/a    & n/a    & 0.50 \\
        SVM   & 0.72    & 0.76    & 0.74 \\
        CNN   & 0.54 & 0.86 & 0.70 \\
        LSTM   & 0.76    & 0.80    & 0.78 \\ \bottomrule
    \end{tabular}%
	\caption{Results excluding sentences that required additional context for manual annotation}
	\label{tab:results_without_context}%
\end{table}%

\begin{table}[htbp]
	\centering
	\begin{tabular}{@{}rccc@{}}
		\toprule
		& $Acc_{\small\hate}$ & $Acc_{\small\nohate}$ & $Acc_{\small\textsc{all}}$ \\ \midrule
		Majority & n/a    & n/a    & 0.50 \\
        SVM   & 0.69    & 0.73    & 0.71 \\
        CNN   & 0.55 & 0.79 & 0.66 \\
        LSTM   & 0.71    & 0.75    & 0.73 \\ \bottomrule
    \end{tabular}%
	\caption{Results including sentences that required additional context for manual annotation}
	\label{tab:results_with_context}%
\end{table}%

\subsection{Error Analysis}
\label{ssec:error}

In order to get a deeper understanding of the performance of the classifiers trained, a manual inspection has been performed on a set of erroneously classified sentences. Two main types of errors have been identified:

\paragraph{Type I errors} Sentences manually annotated as \hate\ but classified as \nohate\ by the system, usually due to a lack of context or to a lack of the necessary world knowledge to understand the meaning of the sentence:
\begin{example}
\exitem \label{it:context} ``Indeed, now they just need to feed themselves, educate themselves, police themselves ad nauseum...`
\exitem \label{it:world} ``If you search around you can probably find ``hoax of the 20th century'' for free on the net.''
\end{example}

In \ref{it:context}, it is not clear without additional context who ``themselves'' are. It actually refers to people of African origin. In its original context, the author was implying that they are not able to feed, police nor educate themselves. This would make the sentence an example of hate speech, but it could also be a harmless comment given the appropriate context. In \ref{it:world}, the lack of world knowledge about what the Holocaust is, or what naming it ``hoax'' implies --i.e., denying its existence--, would make it difficult to understand the sentence as an act of hate speech.

\paragraph{Type II errors} Sentences manually labelled as \nohate\ and automatically classified as \hate, usually due to the use of common offensive vocabulary with non-hateful intent:
\begin{example}
\exitem \label{it:hungarian} ``I dont like reporting people but the last thing I will do is tolerate some stupid pig who claims Hungarians are Tartars.``
\exitem \label{it:crime} ``More black-on-white crime: YouTube - Black Students Attack White Man For Eating Dinner With Black''
\end{example}

In \ref{it:hungarian}, the user accuses and insults a particular individual. Example \ref{it:crime} is providing information on a reported crime. Although vocabulary of targeted groups is used in both cases (i.e., ``Hungarians'', ``Tartars'', ``black''), the sentences do not contain \hate.


\section{Discussion}
\label{sec:disc}

There are several aspects of the introduced dataset, and hate speech annotation in general, that deserve a special remark and discussion.

First, the source of the content used to obtain the resulting dataset is on its own a source of offensive language. Being Stormfront a white supremacists' forum, almost every single comment contains some sort of intrinsic racism and other hints of hate. However, not every expression that contains a racist cue can be directly taken as hate speech. This is a truly subjective debate related to topics such as free speech, tolerance and civics. That is one of the main reasons why this paper carefully describes the annotation criteria for what here counts as hate speech and what not. In any case, despite the efforts to make the annotation guidelines as clear, rational and comprehensive as possible, the annotation process has been admittedly demanding and far from straightforward.

In fact, the annotation guidelines were crafted in several steps, first paying attention to what the literature points about hate speech annotation. After a first round of manual labelling, inconsistencies among the human annotators were discussed and the guidelines and examples were adapted. From those debates we extract some conclusions and pose several open questions. The first annotation criteria (hate speech being a \textit{deliberate attack}) still lacks robustness and a proper definition, becoming ambiguous and subject to different interpretations. A more precise definition of what an \textit{attack} is and what it is not would be necessary: Can an objective fact that however undermines the honour of a group of people be considered an attack? Is the mere use of certain vocabulary (e.g. ``nigger'') automatically considered an attack? With regard to the second annotation criteria (hate speech being \textit{directed towards a specific group of people}), it was controversial among the human annotators as well. Sentences were found that attacked individuals and mentioned the targets' skin colour or religion, political trends, and so on. Some annotators interpreted these as indirect attacks towards the collectivity of people that share the mentioned characteristics.

Another relevant point is the fact that the annotation granularity is sentence level. Most, if not all, of the existing datasets label full comments. A comment might be part of a more elaborated discourse, and not every part may convey hate. It is arguable whether a comment containing a single hate-sentence can be considered ``hateful'' or not. The dataset released provides the full set of sentences per comment with their annotations, so each can decide how to work with it.

In addition, and related to the last point, one of the labels included for the manual labelling is \relation. This label is meant to be used when two or more sentences need each other to be understood as hate speech, usually because one is a premise and the following is the (hateful) conclusion. This label has been seldom used.

\begin {figure}
	\centering
    \resizebox {\columnwidth} {!} {
    \begin{tikzpicture}
		\begin{axis}[
          xlabel=$Annotated\ batches\ of\ 500\ comments\ each$,
          ylabel=$with\ context/all\ items\ ratio$, 
          xmin=1, 
          ymin=0, 
          xmax=10,
          point meta={y*100},
    	] \addplot[smooth,color=blue,mark=.] plot coordinates {
          (1,31.993007)
          (2,16.6512488)
          (3,8.1687612)
          (4,7.7715356)
          (5,3.7391304)
          (6,3.8147139)
          (7,4.8913043)
          (8,3.8235294)
          (9,4.1297935)
          (10,5.3275109)
		};
    	\end{axis}
	\end{tikzpicture}
    }
    \caption{Percentage of comments per batch that required additional context to be manually labelled. The amount of context needed by a human annotator decreases over time.}
    \label{fig:context_usage_plot}
\end {figure}
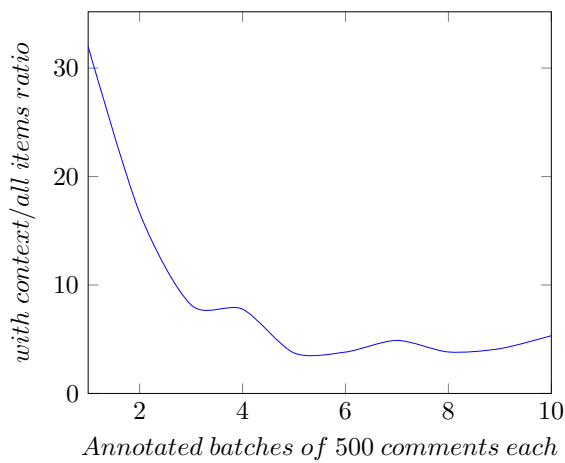

Finally, a very important issue to consider is the need of additional context to label a sentence (i.e., the rest of the conversation or the title of the forum-thread). It can happen to human annotators and, of course, to automatic classifiers, as confirmed in the error analysis (Section \ref{ssec:error}). Studying context dependency to perform the labelling, it has been observed that annotators learn to distinguish hate speech more easily over time, requiring less and less context to make the annotations (see Figure \ref{fig:context_usage_plot}).

\section{Conclusions and Future Work}
\label{sec:conc}

This paper describes a manually labelled hate speech dataset obtained from Stormfront, a white supremacist online forum.

The resulting dataset consists of $\sim$10k sentences labelled as conveying hate speech or not. Since the definition of hate speech has many subtleties, this work includes a detailed explanation of the manual annotation criteria and guidelines. Furthermore, several aspects of the resulting dataset have been studied, such as the necessity of additional context by the annotators to make a decision, or the distribution of the vocabulary used in the examples labelled as hate speech. In addition, several baseline experiments have been conducted using automatic classifiers, with a focus on examples that are difficult for automatic classifiers, such as those that required additional context or world knowledge. The resulting dataset is publicly available.

This dataset provides a good starting point for discussion and further research. As future work, it would be interesting to study how to include world knowledge and/or the context of an online conversation (i.e. previous and following messages, forum thread title, and so on) in order to obtain more robust hate speech automatic classifiers. Future studies could also explore how sentences labelled as \relation\ affect classification, as this sentences have not been included in the experiments presented. In addition, more studies should be performed to characterize the content of the dataset in depth, regarding timelines, user behaviour and hate speech targets, for instance. Finally, since the proportion of \hate/\nohate\ examples tends to be unbalanced, a more sophisticated manually labelling system with active learning paradigms would greatly benefit future labelling efforts.


\section{Acknowledgements}
\label{sec:ack}

This work has been supported by the European Commission under the project ASGARD (700381, H2020-FCT-2015).
We thank the Hatebase team, in particular Hatebase developer Timothy Quinn, for providing Hatebase's English vocabulary dump to conduct this study.
Finally, we would like to thank the reviewers of the paper for their thorough work and valuable suggestions.

\bibliography{emnlp2018}
\bibliographystyle{acl_natbib_nourl}

\end{document}